\documentclass[letterpaper, 10 pt, conference]{ieeeconf}  

\IEEEoverridecommandlockouts                              

\usepackage{graphics} 
\usepackage{amsmath} 
\usepackage{amssymb}  
\usepackage[table]{xcolor}
\usepackage{soul}
\usepackage{graphicx}
\usepackage{makecell}

\usepackage{float}

\usepackage{etoolbox}
\usepackage{booktabs}
\usepackage{hyperref}
\usepackage{subcaption}

\title{\LARGE \bf
Data-Efficient Learning of Natural Language to Linear Temporal Logic Translators for Robot Task Specification
}

\author{Jiayi Pan, Glen Chou, and Dmitry Berenson
\thanks{$^{1}$University of Michigan, Ann Arbor, MI, USA, 48109,
        {\tt\small \{jiayipan, gchou, dmitryb\}@umich.edu}. This work was supported in part by the Office of Naval Research Grant N00014-21-1-2118 and NSF grants IIS-1750489 and IIS-2113401.}%
}

\begin{document}

\newbool{showComments}
\booltrue{showComments}

\ifbool{showComments}{%
\newcommand{\jiayi}[1]{\sethlcolor{yellow}\hl{[Jiayi: #1]}}
\newcommand{\glen}[1]{\sethlcolor{lime}\hl{[Glen: #1]}}
\newcommand{\dmitry}[1]{\sethlcolor{orange}\hl{[Dmitry: #1]}} 

\colorlet{ourcolor}{green!20}
\colorlet{baselinecolor}{orange!20}
\colorlet{ablationcolor}{blue!10}

\newcommand{\ours}[1]{\sethlcolor{ourcolor}\hl{#1}}
\newcommand{\baseline}[1]{\sethlcolor{baselinecolor}\hl{#1}}
\newcommand{\ablation}[1]{\sethlcolor{ablationcolor}\hl{#1}} 

}{
\newcommand{\pjy}[1]{}
\newcommand{\wjl}[1]{}
\newcommand{\ncy}[1]{}
\newcommand{\wf}[1]{}
}
\newcommand{\nicepurple}[1]{{\textcolor{magenta!100}{\fontfamily{qag}#1}}}
\newcommand{\niceblue}[1]{{\textcolor{blue!100}{\fontfamily{qag}#1}}}

\maketitle
\thispagestyle{empty}
\pagestyle{empty}

\begin{abstract}

To make robots accessible to a broad audience, it is critical to endow them with the ability to take universal modes of communication, like commands given in natural language, and extract a concrete desired task specification, defined using a formal language like linear temporal logic (LTL). In this paper, we present a learning-based approach for translating from natural language commands to LTL specifications with very limited human-labeled training data. This is in stark contrast to existing natural-language to LTL translators, which require large human-labeled datasets, often in the form of labeled pairs of LTL formulas and natural language commands, to train the translator. To reduce reliance on human data, our approach generates a large synthetic training dataset through algorithmic generation of LTL formulas, conversion to structured English, and then exploiting the paraphrasing capabilities of modern large language models (LLMs) to synthesize a diverse corpus of natural language commands corresponding to the LTL formulas. We use this generated data to finetune an LLM and apply a constrained decoding procedure at inference time to ensure the returned LTL formula is syntactically correct. We evaluate our approach on three existing LTL/natural language datasets and show that we can translate natural language commands at 75\% accuracy with far less human data ($\le$12 annotations). Moreover, when training on large human-annotated datasets, our method achieves higher test accuracy (95\% on average) than prior work. Finally, we show the translated formulas can be used to plan long-horizon, multi-stage tasks on a 12D quadrotor.

\end{abstract}

\vspace{-5pt}
\section{Introduction}\label{sec:intro}
\vspace{-5pt}

Many tasks that we want our robots to complete are temporally-extended and multi-stage in nature. For example, the success of cooking, urban navigation, robotic assembly, etc. is determined not by a single goal, but rather a sequence of interconnected subtasks and time-varying constraints. Thus, to reliably complete such tasks, it is critical to have an unambiguous specification of these goals and constraints. 

Linear temporal logic (LTL) \cite{DBLP:books/daglib/0020348} is a powerful and expressive tool for unambiguously specifying temporally-extended tasks. LTL augments the traditional notions of standard propositional logic with temporal operators that are able to express properties holding over trajectories; to complete the task, low-level robot trajectories that satisfy the LTL formula can then be synthesized \cite{DBLP:conf/cdc/RamanDMMSS14} to complete the task. Despite its strength in specifying complex tasks, LTL is difficult to use for non-expert end users \cite{DBLP:conf/etfa/PakonenPBV16, DBLP:conf/tacas/SchlorJW98}, and it is unreasonable to expect an end user to provide an LTL formula that encodes the desired task for many applications. In contrast, it is easy for humans to provide natural language commands. Thus, a semantic parser which can translate natural language commands into LTL specifications is of great interest.

\begin{figure}
        \includegraphics[width=\linewidth]{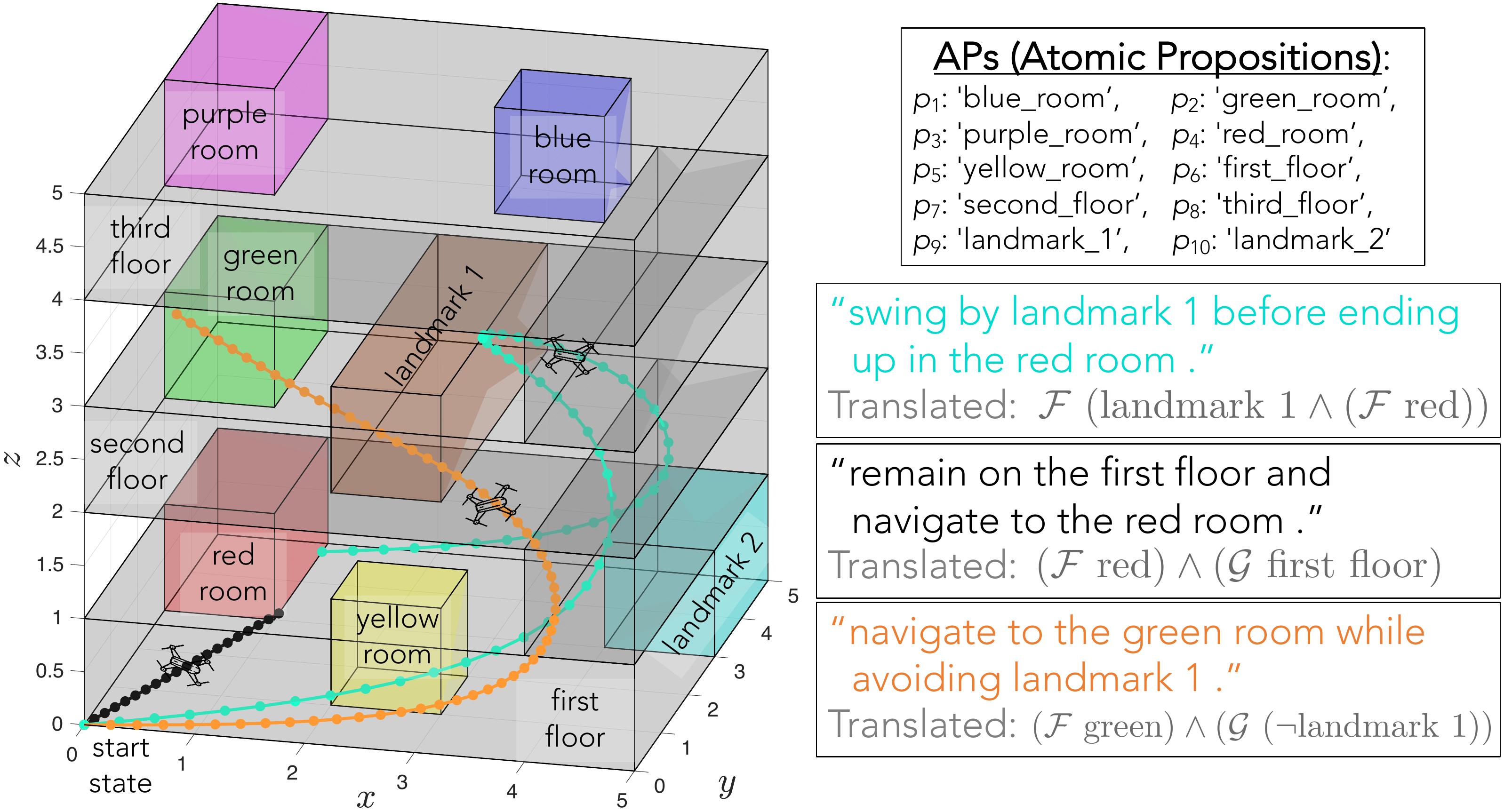}\vspace{-8pt}
    \caption{We translate natural language commands into LTL formulas that achieve complex tasks on a 12D quadrotor.\vspace{-19pt}}
    \label{fig:demo}
\end{figure}

However, training a task-specific semantic parser can be difficult, and requires a large dataset of natural language commands paired with corresponding LTL formulas \cite{oh_planning_2019, gopalan_sequence--sequence_2018}; in particular, to use neural architectures, thousands of annotated examples and hundreds of human workers \cite{oh_planning_2019}, may be required for good generalization. This is prohibitively expensive to collect and is prone to labeling errors, unless LTL experts are used to annotate the data -- hence, obtaining data is the key challenge facing LTL translation. In contrast, recent semantic parsing work in the natural language processing (NLP) community \cite{DBLP:conf/acl/YinWSN22} has alleviated the need for human-annotated data via synthetic training data \cite{xu_autoqa_2020, rongali_training_2022} and the built-in natural language understanding of pre-trained large language models (LLMs) like GPT-3 \cite{DBLP:conf/nips/BrownMRSKDNSSAA20} and BART \cite{DBLP:conf/acl/LewisLGGMLSZ20}. 

In this paper, we reduce the human-labeled data requirements for natural language-to-LTL translators by applying ideas from low-resource semantic parsing. 
We assume we are given a predefined set of possible LTL formulas and atomic propositions, and up to one natural language annotation for each formula.
We translate these pre-defined formulas to (structured) English either by a rule-based translator when the dataset is sufficiently structured, or by querying a human expert for a translation template, and then using the paraphrasing abilities of modern LLMs \cite{DBLP:conf/nips/BrownMRSKDNSSAA20} to generate a large corpus of diverse natural language commands with similar meaning to the associated LTL formulas. We then use this data to finetune an LLM. Here, we explore two variants, where for training labels we use 1) raw LTL formulas, or 2) a canonical form of the LTL formulas \cite{wang_building_2015}) (an intermediate representation between LTL and English). At evaluation time, we enforce the LLM's output to be syntactically consistent with LTL via constrained decoding.
We evaluate our approach on several existing datasets of paired LTL and natural language commands \cite{gopalan_sequence--sequence_2018, oh_planning_2019}, and show our method achieves competitive performance with prior work (trained on thousands of human annotations), with $\le$12 human-labeled annotations. Moreover, when combined with human-labeled data, our method exhibits improved generalization compared to prior work. Overall, our contributions are: 
\begin{itemize}
    \item data augmentation schemes for training natural language-to-LTL translators with very few human annotations,
    \item a neural translation architecture which draws from recent advances in the semantic parsing community to improve LTL translation performance,
    \item evaluation on several datasets in the literature, achieving competitive performance with far fewer human labels.
\end{itemize}

\section{Related Work}\label{sec:related_work}

First, our work is related to methods which aim to obtain task constraints and LTL specifications from human interaction. A large body of work uses interactive training \cite{wang_teaching_2020, DBLP:journals/corr/abs-2003-02232} and physical demonstrations to infer task constraints \cite{DBLP:journals/ijrr/ChouBO21, DBLP:journals/ral/ChouOB20, DBLP:conf/icra/Perez-DArpinoS17, DBLP:conf/iclr/ScobeeS20} and LTL formulas \cite{chou_learning_2022, DBLP:conf/nips/ShahKSL18, chou_explaining_2020, DBLP:conf/nips/Vazquez-Chanlatte18}. However, these forms of human interactions tend to be costly; hence, our goal in this work is to recover LTL formulas from cheaper input, e.g., natural language commands.

If we specify the interaction medium to be language, there is extensive work on translating English to LTL. Early work \cite{finucane_ltlmop_2010, kress-gazit_translating_2008,nikora_automated_2009} translated structured English commands to LTL formulas (possibly through an intermediate structured representation); however, providing English commands with this structure requires an understanding of the specific grammar used, which can be unnatural for humans. More recent work uses neural networks to train the translator using thousands of human-labeled natural language/LTL pairs \cite{oh_planning_2019, gopalan_sequence--sequence_2018, hahn_formal_2022}. To reduce the need for human labels, other work aims to learn from trajectories paired with natural language; this however, still requires many trajectories (i.e., demonstrations or executions) to implicitly supervise the translator  \cite{DBLP:conf/rss/PatelPT20} \cite{wang_learning_2021}. Other work \cite{DBLP:conf/icra/BergBMRPT20} \cite{DBLP:conf/icra/HsiungMCLPT022} improves the translators' generalization to new domains; this is complementary to our method, which improves accuracy within a given set of domains and reduces reliance on human-labeled data. Other work directly translates language to actions \cite{DBLP:conf/cvpr/DasDGLPB18, DBLP:conf/cvpr/AndersonWTB0S0G18} without using LTL, and thus cannot use the planning tools \cite{DBLP:conf/cdc/RamanDMMSS14} that we can exploit.

Our work also relates to the problem of semantic parsing from the NLP community, which seeks to convert from an utterance of (unstructured) natural language to a (structured) logical form which is machine-understandable; e.g., between a command expressed in natural language and an explicit query in a SQL database \cite{sun-etal-2018-semantic}. Recently, significant progress has been made in \textit{low-resource} semantic parsing. Early works in the area \cite{berant-liang-2014-semantic, wang_building_2015} proposed to use a ``canonical" natural language form, i.e., an alternate representation of the formal syntax that is closer to English, and which was shown to improve performance on complex tasks. 
More recent work explores low-resource learning by \cite{xu_autoqa_2020, DBLP:conf/acl/YinWSN22} exploiting automatic training data synthesis using a combination of parsing, templating, paraphrasing, and filtering techniques, or by leveraging large language models (LLMs) \cite{shin_constrained_2021, rongali_training_2022}, such as GPT-3 \cite{DBLP:conf/nips/BrownMRSKDNSSAA20} or BART \cite{DBLP:conf/acl/LewisLGGMLSZ20} for their improved performance and generalization capabilities.
While low-resource semantic parsing is well-studied in NLP, these advances have yet to transfer to natural language-to-LTL translation, which is itself a semantic parsing problem. A key contribution of our work is to bridge the gap between these two communities. Through extensive experiments, we show that recent ideas in low-resource semantic parsing can notably increase the sample efficiency of traditionally data-hungry LTL translators.

\section{Preliminaries and Problem Statement}

We first overview the basics of linear temporal logic (Sec. \ref{sec:prelim_LTL}) and modern generative language models (Sec. \ref{sec:prelim_language}), and then give our problem statement (Sec. \ref{sec:prelim_problem}).

\subsection{Linear temporal logic (LTL)}\label{sec:prelim_LTL}

We consider planning for discrete-time systems $x_{t+1} = f(x_t, u_t)$, with state $x \in \mathcal{X}$ and control $u\in\mathcal{U}$. To specify tasks for this system, we use linear temporal logic (LTL) \cite{DBLP:books/daglib/0020348}, which augments standard propositional logic to express properties holding on system trajectories over periods of time. Similar to \cite{oh_planning_2019}, the LTL specifications considered in this paper can be written with the grammar
\begin{equation}\label{eq:grammar}
	\varphi ::= p \mid \neg p \mid \varphi_1 \wedge \varphi_2 \mid \varphi_1 \vee \varphi_2 \mid \mathcal{G} \varphi \mid \mathcal{F} \varphi \mid \varphi_1\ \mathcal{U}\ \varphi_2,
\end{equation}

\noindent where $p \in \mathcal{P} \doteq \{p_i\}_{i=1}^{N_\textrm{AP}}$ are atomic propositions (APs). In this paper, the APs correspond to sets of salient regions of the state space which the robot may wish to visit or avoid (e.g., the blue room in Fig. \ref{fig:demo} is an AP). As we consider continuous-state systems in this paper, we associate each AP with a constrained region in the state space; that is, $x \models p_i \Leftrightarrow g_i(x) \le 0$, for a constraint function $g_i: \mathcal{X} \rightarrow \mathbb{R}$. Additionally, $\mathcal{G} \varphi$ denotes that the condition $\varphi$ should hold globally for all time, $\mathcal{F} \varphi$ denotes that $\varphi$ should hold eventually (i.e., there exists some time-step $t$ where $\varphi$ is true), and $\varphi_1\ \mathcal{U}\ \varphi_2$ denotes that $\varphi_1$ should hold for all time-steps until $\varphi_2$ holds for the first time. This grammar can be used to specify a diverse set of robotic tasks in, e.g., navigation (``drive to the charging station" as $\mathcal{F}\ p_\textrm{charging}$), manipulation (``empty the mug before stacking" as $\neg p_\textrm{stack} \ \mathcal{U}\ p_\textrm{empty}$), etc.

\subsection{Generative Language Models}\label{sec:prelim_language}
Given a piece of text with $n$ words $w_1, w_2, ..., w_n$, a language model will estimate the probability $p(w_1, w_2, ..., w_n)$, for all possible instantiations of text. Auto-regressive language models factor the probability as 
\begin{equation}\label{eq:autolm}
p(w_1, w_2, ..., w_n)=\prod_{i=1}^n p\left(w_i \mid w_1, \ldots, w_{i-1}\right)
\end{equation}
This formulation of language modeling allows efficient text generation, where given preceding words $w_1, w_2, ..., w_{i-1}$, the model can generate the probability distribution for the next word $p\left(w_i \mid w_1, \ldots, w_{i-1}\right)$. 

Modern transformer-based \cite{NIPS2017_attention} generative language models like GPT-3 \cite{DBLP:conf/nips/BrownMRSKDNSSAA20} and BART \cite{DBLP:conf/acl/LewisLGGMLSZ20} can generate text output in an auto-regressive fashion. They are pre-trained on internet-scale text corpora, and have shown strong natural language understanding and generalization capabilities with impressive performance across many NLP tasks \cite{srivastava2022beyond}.

\subsection{Problem statement}\label{sec:prelim_problem}

In this paper, we wish to learn a natural language-to-LTL translator in a data-efficient manner. Specifically, given:
\begin{enumerate}
    \item a list of possible APs, each with an associated natural language description, e.g., an AP named ``G" has the associated description of ``inside the green room",
    \item a list of possible LTL structures, i.e., a template for an LTL formula with undefined APs, which takes instantiations of those APs as input (this assumption can be relaxed, see Sec. \ref{sec:conclusion}),
\end{enumerate}

\noindent we wish to learn a mapping between natural language and LTL task specifications, i.e., given a natural language command, we aim to translate it to its associated LTL form. 

We consider two data regimes: 1) low-resource scenarios, where we provide limited ($\approx$10) human annotations to train the translator, and 2) the standard data regime, where as in pre-existing language models, we provide thousands of natural language-LTL  pairs for training. In low-resource scenarios, we aim to show our method enables satisfactory translation performance, with only a small performance drop relative to a translator trained on a large set of human annotations. In the standard data regime, we aim to show that our translator architecture improves translation accuracy relative to prior methods trained on the same data.

\begin{figure}
    \centering
    \includegraphics[width=\linewidth]{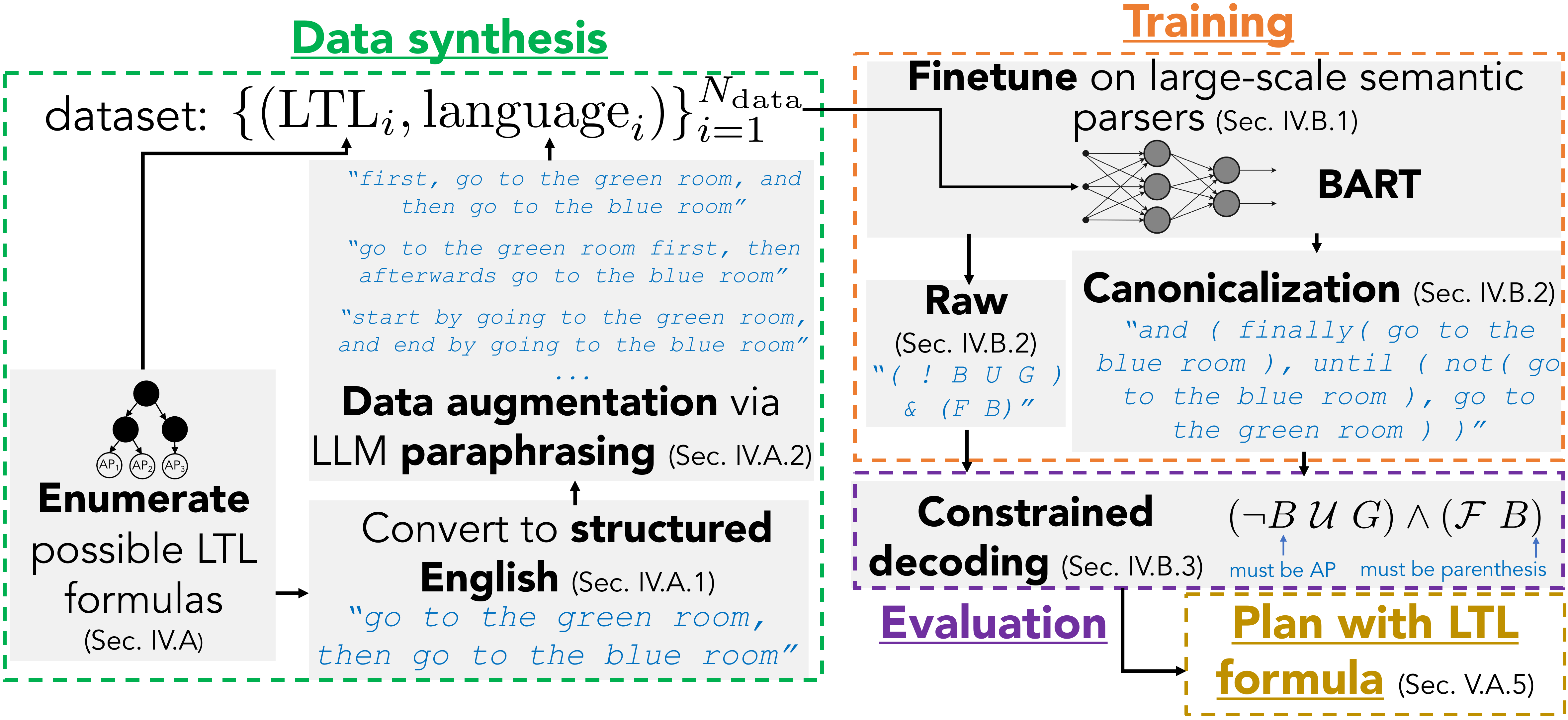}\vspace{-8pt}
    \caption{\small Method flow: generating synthetic data, training on that data, evaluation, and planning with the evaluated formula.\vspace{-20pt}}
    \label{fig:flow}
\end{figure}

\section{Method}
For data-efficent translation of natural language commands to LTL, our method combines 1) a data synthesis pipeline that automatically generates large synthetic training datasets with little human supervision (Sec. \ref{sec:method_data}), and 2) a modern neural semantic parsing architecture that is stronger in natural language understanding and generalization compared to prior work (Sec. \ref{sec:method_arch}). We visualize our method in Fig. \ref{fig:flow}.

\subsection{Data synthesis pipeline}\label{sec:method_data}
Training a neural translator generally requires a corpus of input and output language pairs, e.g., paired natural language commands and LTL formulas as input and output, respectively.
Given the set of possible LTL structures and the set of APs relevant for the set of possible tasks, we can obtain all possible LTL formula outputs by simply filling each LTL structure with combinations of APs.  However, while we can generate large numbers of LTL formulas, obtaining a diverse set of natural language descriptions for each LTL formula typically requires a large amount of human labor, making the training extremely expensive \cite{gopalan_sequence--sequence_2018, oh_planning_2019}.

To alleviate this problem, we apply a two-stage pipeline inspired by \cite{xu_autoqa_2020, DBLP:conf/acl/YinWSN22, wang_building_2015}. First, we perform back-translation (i.e., translate the LTL formula back into \textit{structured} English) and second, perform extensive data augmentation (by leveraging LLMs trained on natural language) to synthesize a diverse set of natural language training data from the LTL formulas, requiring much less human labor. 
During back-translation, given the LTL formulas used in the task, we generate one natural language description for each LTL formula by using either an LTL-to-English translator (when the LTL representation is sufficiently structured), or templates written by human experts (when such structure does not exist). We discuss specific examples of when to use which in Sec. \ref{sec:results}. 
During augmentation, based on the back-translation result, we automatically synthesize a diverse training corpus by leveraging a LLM-based paraphrasing model. We discuss these components in more detail.

\subsubsection{Back-translation} 
Although mapping natural language into a formal language remains an open research question, the inverse problem of mapping formal language back to natural language can be done relatively easily, by either 1) symbolically parsing the formula \cite{ranta2011translating} or 2) training a neural translator \cite{paola-backtranslation}. We build a rule-based LTL-to-English translator based on the grammar of LTL \eqref{eq:grammar}. Given an LTL formula, the translator will parse out its syntax tree and then translate it to structured English. This assumed structure renders the translation straightforward. When the LTL corpus is too complex or ambiguous for the translator to work (as in the datasets explored in Sec. \ref{sec:res_cleanup} and \ref{sec:res_pick}), we obtain the back-translation result by querying human experts to provide a small number of annotations; see Sec. \ref{sec:res_cleanup} and \ref{sec:res_pick} for specific instances of this process. 

\subsubsection{Augmentation}\label{sec:method_data_aug}
Given the training data obtained in back-translation, unlike previous methods \cite{gopalan_sequence--sequence_2018, oh_planning_2019}, which simply augment the dataset by replacing existing AP combinations with novel ones, we follow \cite{xu_autoqa_2020, rongali_training_2022} and use a neural paraphrasing model to \textit{paraphrase} the text. In particular, we prompt the GPT-3 language model \cite{DBLP:conf/nips/BrownMRSKDNSSAA20} to give ten different paraphrases for every English sentence created during back-translation to augment the synthetic training corpus. An example from the data synthesis pipeline in Sec. \ref{sec:res_cleanup} is shown below. This example consists of a prompt template (a text template to be filled with string arguments) filled with \nicepurple{a source natural language command} and then \niceblue{GPT-3's output} as the paraphrased results.

\vspace{-0.2cm}
\begin{center}
\small
\begin{tabular}{p{0.8\columnwidth}}
{Rephrase the source sentence in 10 different ways. Make the outputs as diverse as possible.}\\\\
\end{tabular}

\vspace{-0.2cm}
\begin{tabular}{p{0.8\columnwidth}}
{Source:}
\nicepurple{Go to the blue room or go to the red room to finally go to the yellow room.}\\\\

\vspace{-0.4cm}
{Outputs:}\\
{1.} \niceblue{You can go to the blue room or the red room, and then finally the yellow room.} \\
\niceblue{2. To get to the yellow room, you must go through the blue room or the red room.}\\
\niceblue{...}\\
\niceblue{10. In order to reach the yellow room, you must first go to the blue room or red room.}
\end{tabular}
\end{center}

Since the back-translated structured English commands are empirically similar to the natural language that GPT-3 is trained on, we find GPT-3 returns meaningful, diverse paraphrases resembling natural language.
In short, our insight is to exploit LLMs' large-scale pre-training on general-purpose natural language to generate diverse English commands that notably reduces reliance on human annotators (who may also make mistakes due to unfamiliarity with LTL, cf. Sec. \ref{sec:intro}).

\subsection{Architecture}\label{sec:method_arch}
Applying large language models to low-resource semantic parsing has led to much progress (see Sec. \ref{sec:related_work}).
Following \cite{shin_constrained_2021}, we select the pre-trained BART-large language model as our translation model, finetune it on the task-specific corpus (using either raw LTL or a canonicalization of LTL for training labels), and at inference time perform LTL grammar-constrained decoding. We discuss these now in detail.

\subsubsection{Pre-training and fine-tuning}
BART is a transformer-based \cite{NIPS2017_attention} language model. Given a corrupted version of English text as input, e.g., ``my is Alex", the model is trained to recover and output the original text `my name is Alex". In our context, BART is given the natural language command as input, and the LTL formula as output. We explore two variants on the training label representation: 1) using the raw LTL formula for training labels (i.e., $\mathcal{F}\ B$, for ``eventually visit the blue room", is transcribed as ``F B" for the training label), and 2) using a canonicalization of the LTL formula for the labels (an intermediary between LTL and English), which we describe in Sec. \ref{sec:method_canonical}.
It is worth noting that our proposed method can be easily applied to other potentially stronger language models like T5-XXL \cite{2020t5} or GPT-3 \cite{DBLP:conf/nips/BrownMRSKDNSSAA20}; we choose BART-large because it has a moderate number of 406M parameters and is efficient to finetune on a single GPU. We use the hyper-parameters from \cite{shin_constrained_2021} for finetuning.

\subsubsection{Canonical form for LTL} \label{sec:method_canonical}
While exploiting the structure in pre-trained LLMs can be fruitful, directly applying them on LTL formulas (especially) can degrade performance. As language models (including BART) are primarily trained on natural language, there is a distribution shift when evaluating on the text transcription of LTL formulas, e.g., ``F B" does not resemble natural language.
In \cite{DBLP:conf/acl/LewisLGGMLSZ20}, it was shown that creating a one-to-one mapping from a formal language to a ``canonical" representation, which is ``closer" to natural language than the raw LTL formula, can mitigate the distribution shift and enable stronger benefits from pre-trained LLMs.

\begin{figure}
    \centering
    \includegraphics[width=\linewidth]{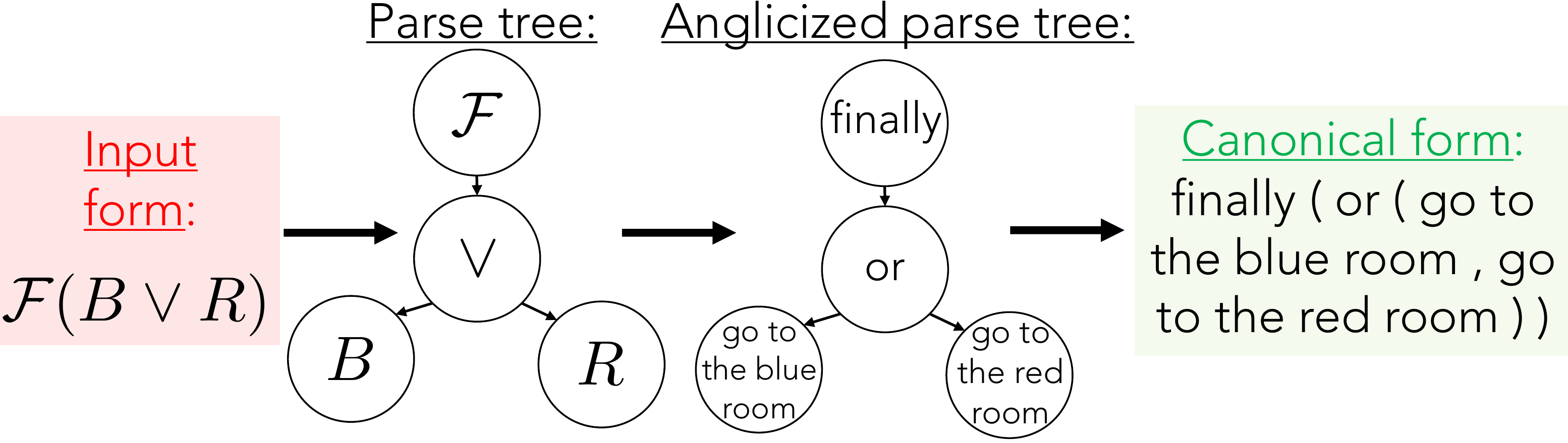}\vspace{-12pt}
    \caption{Transforming from raw LTL to a canonical form.\vspace{-20pt}}
    \label{fig:parse}
\end{figure}

We now describe the canonical form for LTL that we use. Given an LTL formula, we build its equivalent parse tree form (see Fig. \ref{fig:parse}, and \cite{chou_explaining_2020} for details), replace the elements of the LTL grammar with corresponding English phrases, and starting from the parse tree's root, we transcribe it to text, with parentheses and commas to encompass and separate an operator's input arguments. For example (Fig. \ref{fig:parse}), consider the formula $\mathcal{F}(B \vee R)$; this can be written as the parse tree in Fig. \ref{fig:parse}, and after Anglicization and transcription, we have ``finally\ (\ or (\ go to the blue room , go to the red room\ )\ )". 

However, canonicalization also has drawbacks, e.g., 1) it increases the transcription length, which can hurt accuracy, and 2) for simpler tasks, the inductive bias provided by the canonical form may not help as much. Thus, we evaluate both raw and canonicalized LTL in the results to explore which representation is better suited for LTL translation.

\subsubsection{Constrained decoding for the language model}
Constrained decoding \cite{shin_constrained_2021} is a common technique used together with LLMs in low-resource semantic parsing to guarantee that the output will be well-formed. Given a pre-defined set of possible outputs, the system will constrain the LM by only considering the next-token prediction that is in the output sets. 
In practice, we incorporate the constrained decoding implementation in \cite{shin_constrained_2021} and provide it with the set of possible LTL formulas in the task obtained in \ref{sec:method_data}.

To recap, we synthesize a dataset of natural language/LTL pairs by generating possible LTL formulas, converting them to structured English, and then using LLM paraphrasers to get synthetic natural language commands. This data (either in raw or canonical form) is used to finetune BART, and at evaluation time, we use an LTL-constrained decoder.

\vspace{-3pt}
\section{Results}\label{sec:results}
\vspace{-3pt}
To evaluate our approach, we compare our method (the raw LTL and canonical variants are denoted as BART-FT-Raw and BART-FT-Canonical in  in Tab. \ref{tab:drone}) with two existing baselines for natural language-to-LTL translation: CopyNet \cite{DBLP:conf/icra/BergBMRPT20}, and an RNN with attention mechanism\footnote{Two variants of RNN models are discussed in \cite{gopalan_sequence--sequence_2018}, which have very similar performance. We select the RNN + Bahdanau Attention architecture \cite{Bahdanau2015NeuralMT} for our experiments as it has overall better performance.} (denoted RNN) \cite{gopalan_sequence--sequence_2018}. 
 We also examine several ablations of our method, to evaluate the necessity of various components of our pipeline. In particular, we 1) remove constrained decoding at evaluation time, denoted ``-NoConstrainedDecoding", and 2) train BART directly on structured English, without paraphrasing (cf. Sec. \ref{sec:method_data_aug}), denoted ``no augmentation" in Tab. \ref{tab:drone}.

We evaluate our method on three datasets of paired LTL formulas and natural language commands: a drone planning dataset (Sec. \ref{sec:res_drone}) \cite{oh_planning_2019}, an robot navigation dataset (Sec. \ref{sec:res_cleanup}) \cite{gopalan_sequence--sequence_2018}, and a robot manipulation dataset (Sec. \ref{sec:res_pick}) \cite{gopalan_sequence--sequence_2018}. We show that 1) despite our limited human-labeled data, we achieve competitive English to LTL translation accuracy on these datasets, and 2) when trained on the datasets, our architecture yields better accuracy than the baselines. Our code is at \nicepurple{\href{https://github.com/UM-ARM-Lab/Efficient-Eng-2-LTL}{github.com/UM-ARM-Lab/Efficient-Eng-2-LTL}}.

\begin{figure}
    \centering
    \includegraphics[width=\linewidth]{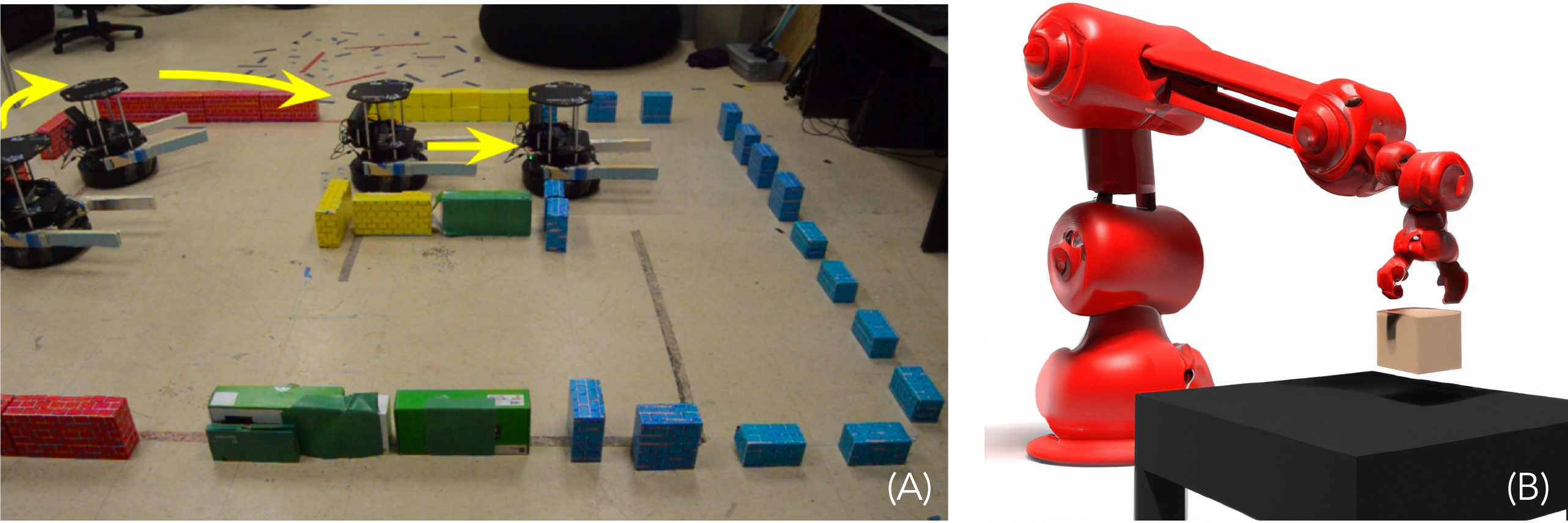}
    \caption{The evaluation datasets. See Fig. \ref{fig:demo} for the drone dataset. (A) Cleanup World \cite{gopalan_sequence--sequence_2018}. (B) Pick-and-place \cite{gopalan_sequence--sequence_2018}.\vspace{-20pt}}
    \label{fig:visual}
\end{figure}

\vspace{-6pt}
\subsection{Drone planning }\label{sec:res_drone}
\vspace{-3pt}
\subsubsection{Definition}
In this dataset (from \cite{oh_planning_2019}), as illustrated in Fig. \ref{fig:demo}, the task is to translate a natural language command for drone navigation into an LTL expression, which can then be fed into a trajectory planner that completes the task in a pre-defined environment (i.e., if the correspondence between an AP and its real-world region is known). This dataset contains 5 unique LTL structures and 12 different APs, with a total of 6,185 commands for 343 different LTL formulas. 

\subsubsection{Experimental setup}
To explain the structure and our processing of this dataset, we present an example below. In black, we show the instruction in natural language, followed by \nicepurple{the canonical form} (see Sec. \ref{sec:method_canonical}) used by our method, and \niceblue{the raw LTL representation} used by the baseline: 

\vspace{-0.2cm}
\begin{center}
\small
\begin{tabular}{p{0.8\columnwidth}}
{head to the yellow room , but make sure to go through the blue room first .}\\ \hline
\nicepurple{finally ( and ( the blue room , finally ( the yellow room ) )}\\ \hline
\niceblue{F ( blue\_room \& F ( yellow\_room ) )}
\end{tabular}
\end{center}
\vspace{-0.2cm}

Of the three considered datasets, the LTL formulas in this dataset are neither too ambiguous nor too complex (see Sec. \ref{sec:res_cleanup} and \ref{sec:res_pick} respectively for cases where it does not hold) to stop back-translation from functioning. Thus, we first map each original LTL representation to its canonical form via parse tree (cf. Sec. \ref{sec:method_canonical}), and then do the back-translation.

\subsubsection{Results}
\begin{table*}[ht!]
\begin{center}
\caption{\footnotesize Translation accuracy. \protect\ours{Ours}, \protect\baseline{baselines}, \protect\ablation{ablations}. Top: regular data regime; bottom: low-resource regime. (Number of LTL structures/formulas).\vspace{-5pt}}
\begin{tabular}{l l l c c c c c c}
\hline
Model architecture & Training data & Test data & Drone (5/343) & Cleanup (4/39) & Pick (1/5)\\
\hline
\rowcolor{orange!20}RNN \cite{gopalan_sequence--sequence_2018} & 4/5 golden & 1/5 golden & 87.18 & 95.51 & 93.78\\
\rowcolor{orange!20}CopyNet \cite{DBLP:conf/icra/BergBMRPT20} & 4/5 golden & 1/5 golden & 88.97 & 95.47 & 93.14 \\
\rowcolor{green!20}BART-FT-Raw (ours) & 4/5 golden & 1/5 golden & \bf{90.78} & \bf{97.84}  & \bf{95.97}\\
\rowcolor{green!20}BART-FT-Canonical (ours) & 4/5 golden & 1/5 golden & {90.56} & {97.81}  & {95.70}\\ \Xhline{5\arrayrulewidth}
\rowcolor{orange!20}RNN \cite{gopalan_sequence--sequence_2018}& synthetic & full golden & 22.41 & 52.54 & 32.39 \\
\rowcolor{orange!20}CopyNet \cite{DBLP:conf/icra/BergBMRPT20} & synthetic & full golden & 36.41 & 53.40 & 40.36\\
\rowcolor{green!20}BART-FT-Raw (ours) & synthetic & full golden & \bf{69.39}  & \bf{78.00} & \bf{81.45} \\ 
\rowcolor{green!20}BART-FT-Canonical (ours) & synthetic & full golden & {68.99} & {77.90} & {78.23} \\ 
\rowcolor{blue!10}BART-FT-Raw-NoConstrainedDecoding & synthetic & full golden & 68.23  & 76.26 & 81.05 \\ 
\rowcolor{blue!10}BART-FT-Canonical-NoConstrainedDecoding & synthetic & full golden & 67.45  & 72.06 & 69.49 \\ 
\rowcolor{blue!10}BART-FT-Raw (ours) & synthetic; no augmentation & full golden & 29.43 & 52.51 & 80.38\\ 
\rowcolor{blue!10}BART-FT-Canonical (ours) & synthetic; no augmentation & full golden & 39.21 & 53.16 & 67.88\\ 
\hline
\end{tabular}\vspace{-24pt}
\label{tab:drone}
\end{center}
\end{table*}

Our translation accuracy on this dataset is presented in Tab. \ref{tab:drone}. The translation output is considered accurate if it matches exactly with the ground truth output. This may be conservative, since some clauses in a formula can be reordered (thus failing to match the output exactly) while retaining identical semantic meaning (see Sec. \ref{sec:conclusion} for more discussion). In the training data column of Tab. \ref{tab:drone}, ``golden" refers to the human-annotated data from the original drone planning dataset, while ``synthetic" refers to the synthetic training data that we obtained by the data synthesis pipeline of Sec. \ref{sec:method_data}. As there is no official division of the training / evaluation split when evaluating on the golden dataset, we report accuracy by its five-fold cross-validation result. We generate 5900 synthetic
 data points and as no golden data is provided to the model for training, we evaluate the model's performance on the full golden dataset.

\subsubsection{Discussion}
When using the golden dataset to train, our model performs the best compared to the baseline models, outperforming them in translation accuracy by about 2\%. Moreover, our ``-Raw" and ``-Canonical" variants have similar accuracy. This suggests  our architecture has better generalization to unseen data, which can be attributed to 1) our model's higher capacity relative to the baselines, and 2) the extensive pre-training provided by BART (in contrast, only the word embedding layer in the baselines is pre-trained). When we consider the low-resource scenario, our method achieves an accuracy of 69\%. Note that 1) reduced accuracy compared to training on the golden dataset is expected, due to the distribution shift between the two datasets, and 2) while application-dependent, accuracies of 70\% are common for the state-of-the-art in semantic parsing, e.g., \cite{shin_constrained_2021}. In contrast, all the baseline methods perform much worse (20-30 \%). The ablation of our method without data augmentation does similarly poorly (20-30\%), whereas removing constrained decoding causes a slight degradation of 1\%. Here, canonicalization hurts performance by 1\%; this may be due to the reasons discussed in Sec. \ref{sec:method_canonical}. On this dataset, we posit that the combination of the pre-trained LLM, the data augmentation, and constrained decoding enables our accuracy, while canonicalization is not needed.

\subsubsection{From LTL formulas to trajectories}
To show that our translated LTL formulas can specify the complex behavior requested in natural language, we compute plans satisfying translated formulas on a quadrotor. It is modeled as a 12D double integrator, where the state is the 3D pose (6 states) and the linear/angular velocity (6 states); we assume we control the accelerations. These are linear dynamics, so dynamically-feasible trajectories satisfying the LTL formulas can be computed with mixed integer convex programming \cite{chou_explaining_2020, DBLP:conf/cdc/RamanDMMSS14}. In Fig. \ref{fig:demo}, we visualize three plans which satisfy the translated formulas. Here, APs are modeled as polytopes, i.e., $p_i \Leftrightarrow \{x \mid A_i x \le b_i\}$. Complex behavior arises from the plans, e.g., for the command ``swing by landmark 1 before ending up in the red room", the drone visits the second floor without exiting the map (gray), touches landmark 1, and then smoothly returns to the first floor to visit the red room.

\vspace{-5pt}
\subsection{Cleanup World}\label{sec:res_cleanup}
\subsubsection{Definition} 
The Cleanup World environment \cite{MacGlashan2015GroundingEC} (Fig. \ref{fig:visual}(A)) involves a robot interacting with its environment by moving through different rooms, or by moving objects from one room to another.
Based on this environment, \cite{gopalan_sequence--sequence_2018} collects 3,382 natural language command-LTL pairs, containing 39 LTL formulas with 4 unique LTL structures and 6 unique APs. The task for this domain is to give a natural language command to a simulated robot, which asks it to move through different rooms or asks to move objects to other rooms. 

\subsubsection{Experiment setup} \label{sec:res_cleanup_exp}
As done for the drone dataset, we will present an example of the structure and our processing of this dataset. The color-coding is the same as in Sec. \ref{sec:res_drone}, but this time we show two instructions in natural language that correspond to the same LTL formula. 

\vspace{-0.2cm}
\begin{center}
\small
\begin{tabular}{p{0.8\columnwidth}}
{enter the red room and bring the chair back to the blue room}\\
\hline
{move into the red room and push the chair back into the purple room}\\ \hline
\nicepurple{finally ( and (  go to the red room , finally ( go to the blue room with chair )  )  )}\\ \hline
\niceblue{F \& R F X}
\end{tabular}
\end{center}
\vspace{-0.2cm}

This dataset lacks documentation for some APs, i.e., it is unclear what ``X" corresponds to in English; without this information, back-translation to structured English cannot be done via our rule-based translator. Moreover, the dataset is highly noisy, e.g., in the second natural language command, the annotator misjudged the color as purple. To handle these challenges, we manually inspect the dataset, and provide the data needed to pair every LTL formula in this domain to a corresponding canonical form/natural language description. Na\"ively, this requires 39 annotations (one for each LTL formula in the dataset), but we reduce this to 10 annotations by exploiting the compositional structure of LTL. Specifically, we collect one natural language description for each of the six APs, and the canonical form/natural language description for each of the four LTL structures. 
It is worth discussing the comparability of data collection costs. Providing a natural language description for the four LTL \textit{structures} may require the human annotator to be more familiar with LTL, while annotating LTL formulas case by case may be easier (has been done with crowd-sourcing \cite{gopalan_sequence--sequence_2018}, though accuracy is still a challenge). Since our pipeline is flexible, one can choose between 10 natural language annotations on LTL structures (more expensive) or 39 cheaper annotations of LTL formulas.

\subsubsection{Results and Discussion}

We report our accuracy in Tab. \ref{tab:drone}. The evaluation criteria (exact matching) is the same.
When using the golden dataset to train the model, like before, both the raw and canonical variants of our method outperform the baselines by 2\%. In the low-resource scenario, we generate 594 synthetic data points, and our method achieves $\approx$78\% accuracy (for both raw and canonical); this is higher than the drone example, and is a $\approx$20\% drop from training on the golden dataset (expected due to distribution shift). In contrast, all baselines perform much worse ($\approx$50\%). The ablations also degrade ($\approx$74\% when removing constrained decoding, $\approx$53\% when removing augmentation), and the ablated raw and canonical variants perform similarly. Overall, this corroborates the conclusions of Sec. \ref{sec:res_drone}.

\vspace{-5pt}
\subsection{Pick-and-place}\label{sec:res_pick}
\vspace{-2pt}
\subsubsection{Definition} In this dataset \cite{gopalan_sequence--sequence_2018} (see Fig. \ref{fig:visual}(B)), the robot conducts repetitive actions based on a user command specified in natural language. It has 5 different LTL formulas with 5 different APs and 1 unique LTL structure.

\subsubsection{Experiment setup}
As done for the previous datasets, we will present an example of the structure and our processing of this dataset; color-coding is as before.

\vspace{-0.2cm}
\begin{center}
\small
\begin{tabular}{p{0.8\columnwidth}}
{scan the empty area of the table and pick up any non green objects moving them to the basket}\\ \hline
\nicepurple{globally ( and ( until ( scan , not ( any non green cubes ) ) , finally ( any non green cubes ) ) )}\\ \hline
\niceblue{G \& U S ! C F C}
\end{tabular}
\end{center}
\vspace{-0.2cm}

As the LTL structure has a parse tree of depth 5 (i.e., the task is complex), it would require extensive engineering to design the LTL-to-English translator. Thus, we follow the process in Sec. \ref{sec:res_cleanup_exp}, and manually inspect the dataset, giving a total of 5 canonical form/natural language annotations. 

\subsubsection{Results and Discussion}\label{sec:res_pick_disc}

Our accuracy on the pick-and-place dataset is presented in Tab. \ref{tab:drone}. The evaluation criteria (exact matching) is the same as before.
Again, when training on the golden dataset, our model (both variants) outperforms the baselines by $\approx$2\%. For the low-resource scenario, we generate 55 synthetic data points, and our method (raw) gives an accuracy of 81\%; this is comparable with Sec. \ref{sec:res_cleanup}, and is around a 14\% drop from training on the golden dataset, which is a slightly smaller drop compared to the other two datasets. Here, canonicalization hurts accuracy by 3\%; this is consistent with Sec. \ref{sec:res_drone}. In contrast, all the baselines perform much worse (32 and 40 \%). The ablations of our method also worsen ($\approx$80\% for the raw variants and $\approx$68\% for the canonical variants). Surprisingly, ``-Raw" degrades less than ``-Canonical" (drop of 1 vs. 13\%). This may be since: 1) there is only one LTL structure, so only the APs need to be correctly translated for overall correctness, and 2) raw LTL is more compact than the canonical form, so there are fewer words to distract the model in identifying the APs. 

Overall, these results are as expected. However, we did not expect ``-Raw" to consistently outperform ``-Canonical", in contrast to established results, e.g., \cite{berant-liang-2014-semantic, wang_building_2015, shin_constrained_2021}. We believe that the most likely reason for this (see Sec. \ref{sec:method_canonical} for other ideas) is that our evaluation datasets are not complex enough to benefit from canonicalization. This is consistent with how the accuracy gap for pick-and-place is smaller than the gap for e.g., the more complex drone dataset.

\section{Discussion and Conclusion}\label{sec:conclusion}

In this paper, we present an approach for translating natural language commands into corresponding LTL formulas. Our method is highly data-efficient, and can achieve 75\% translation accuracy with only a handful of ($\le 12$) human annotations. We achieve this efficiency through data augmentation and by using this data to finetune an LLM.

Our work has limitations that are interesting directions for future work. First, exploiting the language models' uncertainty (e.g., the top $k$ best formulas) by grounding them to the environment may improve accuracy. Second, we assume a natural language command maps to one LTL formula; however, many natural language commands are inherently ambiguous. Thus, we will study uncertainty-aware planning (e.g., \cite{DBLP:conf/corl/ChouBO20, DBLP:journals/ral/ChouWB22}) at the task level, with uncertainty driven by natural language. Third, we assumed we know all possible LTL structures; we will explore automatic synthesis of LTL structures to improve accuracy on unseen LTL structures.








\bibliographystyle{IEEEtran}
\bibliography{references}
\end{document}